\documentclass[letterpaper, 10 pt, conference]{ieeeconf}  % Comment this line out if you need a4paper

\IEEEoverridecommandlockouts                              % This command is only needed if 
\usepackage{graphicx}
\usepackage{amsmath,amssymb,amsfonts}
\usepackage{algorithmic}
\usepackage{algorithm}
\usepackage{xcolor}
\usepackage{animate}
\usepackage{subfig}
\usepackage{tabularx,booktabs}
\usepackage{url}

\title{\LARGE \bf
Deep Fusion of Ultra-Low-Resolution Thermal Camera and Gyroscope Data for Lighting-Robust and Compute-Efficient Rotational Odometry %Multi-Variate Support Vector PID Controller
}

\author{Farida Mohsen$^{1}$, Ali Safa$^{1}$% <-this % stops a space
\thanks{$^{1}$ College of Science and Engineering, Hamad Bin Khalifa University, Doha, Qatar}%
\thanks{Ali Safa supervised the project as Principal Investigator. All authors contributed to the writing of the manuscript.
        {\tt\small asafa@hbku.edu.qa}}%
}

\begin{document}

\maketitle
\thispagestyle{empty}
\pagestyle{empty}

%%%%%%%%%%%%%%%%%%%%%%%%%%%%%%%%%%%%%%%%%%%%%%%%%%%%%%%%%%%%%%%%%%%%%%%%%%%%%%%%

\begin{abstract}  

Accurate rotational odometry is crucial for autonomous robotic systems, particularly for small, power-constrained platforms such as drones and mobile robots. This study introduces thermal-gyro fusion, a novel sensor fusion approach that integrates ultra-low-resolution thermal imaging with gyroscope readings for rotational odometry. Unlike RGB cameras, thermal imaging is invariant to lighting conditions and, when fused with gyroscopic data, mitigates drift which is a common limitation of inertial sensors. We first develop a multimodal data acquisition system to collect synchronized thermal and gyroscope data, along with rotational speed labels, across diverse environments. Subsequently, we design and train a lightweight Convolutional Neural Network (CNN) that fuses both modalities for rotational speed estimation. Our analysis demonstrates that thermal-gyro fusion enables a significant reduction in thermal camera resolution without significantly compromising accuracy, thereby improving computational efficiency and memory utilization. These advantages make our approach well-suited for real-time deployment in resource-constrained robotic systems. Finally, to facilitate further research, we publicly release our dataset as supplementary material. %at \texttt{https://tinyurl.com/y385prj4}.

\end{abstract}

\section*{Supplementary Material}

The dataset used in this work is openly available at: 
\\
\texttt{https://tinyurl.com/y385prj4}

\section{Introduction}

Accurate, energy-efficient, and robust odometry is fundamental to autonomous robotic systems, enabling precise navigation for drones, rovers, and mobile robots. Traditionally, Inertial Measurement Units (IMUs), which integrate an accelerometer, gyroscope, and magnetometer, have been widely used for odometry \cite{imucalibr}. IMUs typically operate at high frequencies (100–1,000 Hz), making them well-suited for capturing highly dynamic motion \cite{huai2022robocentric}. However, IMU-based odometry suffers from drift over time due to cumulative integration errors and sensor biases, which limit its long-term reliability \cite{odometrysurvey, imubias, Leutenegger2015}. To address this issue, IMUs are often integrated with visual data, such as RGB camera inputs, to correct accumulated inertial navigation errors, forming a visual-inertial odometry (VIO) system \cite{huai2022robocentric, visualinertial}.

VIO systems have been successfully applied in robotic navigation \cite{navodom1, navodom2} and Simultaneous Localization and Mapping (SLAM) frameworks \cite{slamay, slamscar}. However, a major limitation of RGB cameras is their dependence on lighting conditions, which can degrade odometry accuracy in low-light and nighttime environments \cite{sensorfusion, navodom2}. To improve the robustness of VIO systems under varying illumination, researchers have explored the integration of alternative sensing modalities such as radar, LIDAR, event-based cameras, high-dynamic-range (HDR) cameras, and high-resolution thermal cameras \cite{sensorfusion, droneradar, navodom2, thermalhigh}.

\begin{figure}[t]
\centering
    \includegraphics[scale = 0.235]{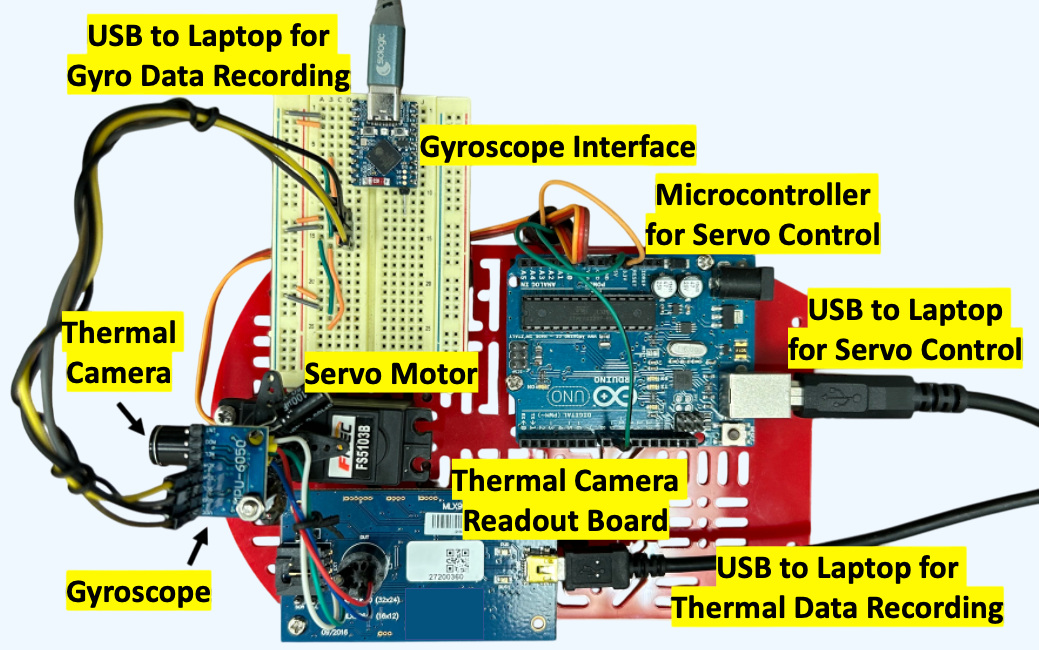}
    \caption{\textit{\textbf{Data acquisition setup.} The $24\times 32$ thermal camera is connected to a readout board which translates its I2C interface to a serial interface via USB. A $100 \mu$F decoupling capacitor is used for providing a stable power supply to the thermal camera. The thermal camera is mounted on top of a servo motor controlled by a micro-controller via serial interface over USB. This setup enables the acquisition of thermal camera data while rotating the camera at precisely controlled speeds.}}
    \label{recordingsetup}
    \vspace{-20pt}
\end{figure}

Among these, ultra-low-resolution thermal cameras offer key advantages in terms of lighting invariance, low power consumption, and compact form factor \cite{Zhao2020, thermalhigh, multimodaldata}. In contrast, radar systems are power-intensive, requiring multiple antennas with power amplifiers to achieve sufficient transmission power at high frequencies (e.g., 79 GHz is a commonly used frequency) \cite{droneradar}. LIDAR sensors remain bulky \cite{sensorfusion}, while HDR and event-based cameras, despite their advantages, are costly and still susceptible to complete darkness \cite{hdrcam, dvssurvey}. Although high-resolution thermal cameras offer much better feature details compared to low-resolution ones, their adoption remains limited due to their bulky form factor and high cost, often priced around $1000$ \$ a unit  \cite{thermalgesture, lowresthermal}.

%To minimize the cost of thermal-based odometry systems, this work introduces the first demonstration of rotational odometry utilizing an ultra-low-resolution (24×32) thermal camera, significantly reducing sensor costs. 
In this work, we are interested in studying the use of \textit{ultra-low-resolution} thermal cameras (e.g., using $24\times 32$ pixels and costing around $30$ \$ a unit) as an attractive solution for the design of cost-efficient yet environmentally-robust odometry systems. However, achieving accurate odometry with ultra-low-resolution thermal cameras remains challenging due to the degradation in visual features that their low resolution leads to, jeopardizing efficient feature tracking across frames.

\begin{figure*}[htbp]
%\vspace{-10pt}
\centering
    \includegraphics[scale = 0.55]{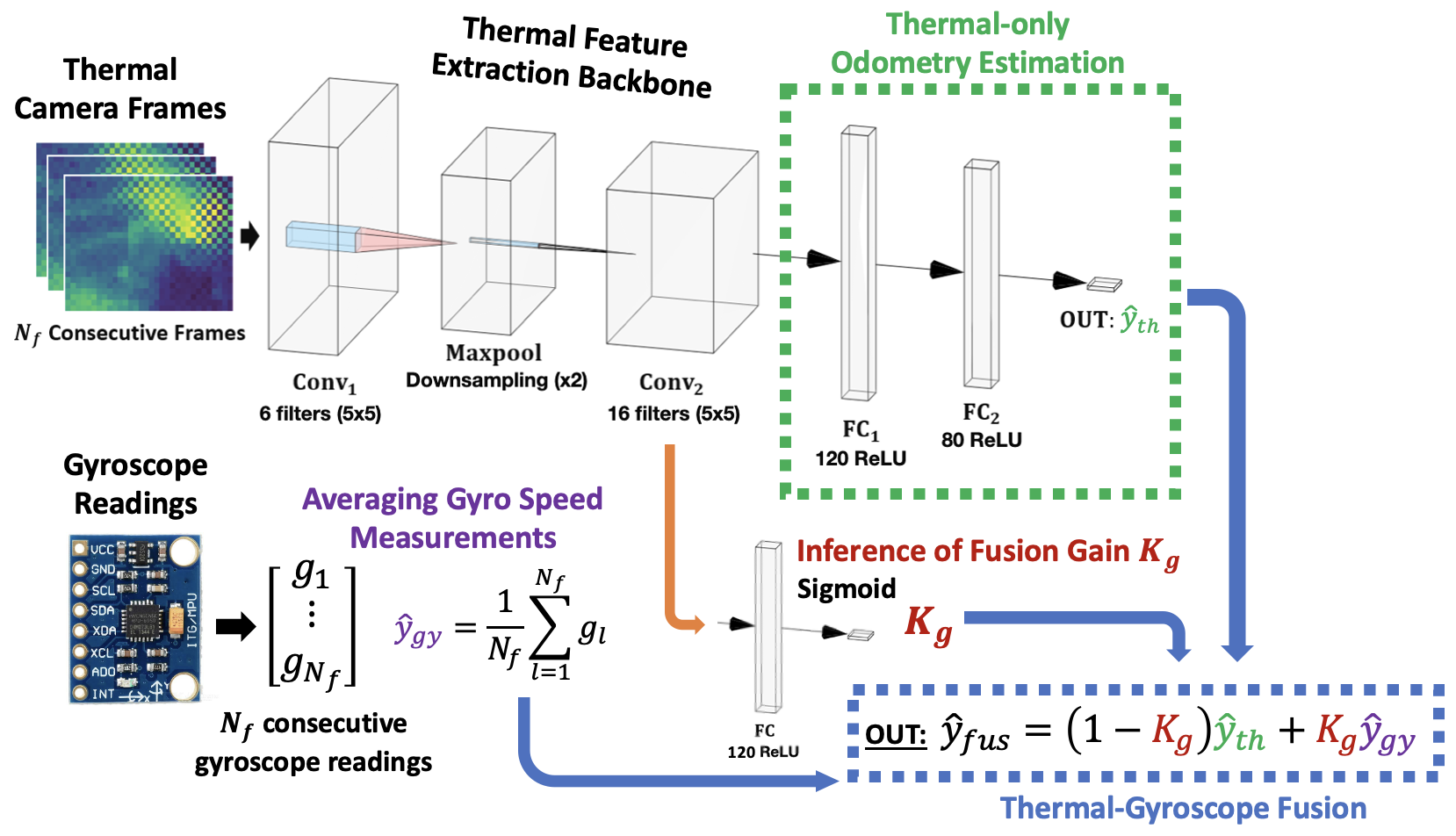}
    \caption{\textit{\textbf{CNN fusion architecture for rotational speed estimation using thermal and gyroscope data.} The network consists of a thermal feature extraction backbone with two convolutional layers and an intermediate max pooling layer, followed by two fully connected layers and a linear output layer for thermal-only odometry estimation. A separate module infers the fusion gain parameter \( K_g \) using a fully connected layer with a sigmoid activation function. The final thermal-gyroscope fusion block adaptively combines the thermal-based and gyroscope-based rotational speed estimates using \( K_g \), balancing the contribution of both modalities based on input data quality.}}
        \label{cnnarch}
    \vspace{-10pt}
\end{figure*}

Building on these observations, this study introduces thermal-gyro fusion, a novel approach that integrates ultra-low-resolution thermal imaging with gyroscope readings for rotational odometry estimation. This fusion compensates for the limited spatial detail in ultra-low-resolution thermal images while mitigating IMU drift, a persistent issue in inertial sensor-based odometry. Additionally, it enables the use of even lower resolution thermal inputs without sacrificing accuracy, significantly reducing computational complexity and power consumption. The contributions of this paper are as follows:
\begin{itemize}
    \item We develop a custom multi-modal data acquisition system, integrating an ultra-low-resolution (24×32) thermal camera and a gyroscope to collect synchronized thermal and gyroscope motion data with precisely labeled rotational speed across diverse environments (see Fig. \ref{recordingsetup}).

   \item We introduce a lightweight CNN-based model designed to fuse thermal imaging with gyroscopic data, leveraging the complementary strengths of both modalities for estimating the rotational speed from both data sources.

   \item We compared the performance of thermal-gyro fusion to a thermal-only configuration on rotational speed estimation. Additionally, we investigate the influence of thermal image resolution and the number of consecutive input thermal frames, demonstrating that fusion enables further resolution reduction without compromising precision.

   \item We publicly release our novel multimodal dataset to support future research in thermal camera-based odometry.
   
\end{itemize}

This paper is structured as follows. Section \ref{dataacq} describes the data acquisition system and dataset. Section \ref{cnnsetup} details the CNN architecture and training methodology. Section \ref{resultssection} presents experimental results and ablation studies. Finally, Section \ref{concsection} concludes the paper.

\section{Data Acquisition}
\label{dataacq}
To acquire \textit{labelled} datasets of thermal camera data and gyroscope readings together with their azimuth rotational speed, the data acquisition setup illustrated in Fig. \ref{recordingsetup} has been constructed. This setup is managed through a Python script on an external laptop, which controls the thermal camera and gyroscope module at different rotation speeds. Meanwhile, it simultaneously records both the thermal camera and gyro data, creating a labeled dataset $\{\Tilde{X}, y\}$ where $\Tilde{X}$ is a sequence of joint thermal and gyro data associated with a rotation speed $y$.

The data set outlined in Table \ref{datasettable} was collected in various indoor and outdoor environments. During data acquisition, the camera’s rotational speed was varied between $20$ deg/s and  $200$ deg/s in both positive and negative directions, with a fixed frame rate of $8$ fps. This approach resulted in a comprehensive dataset comprising $51,561$ thermal camera frames and corresponding gyroscope readings, covering 50 distinct rotation speeds across 18 different environmental settings.

\begin{table}[htbp]
\centering
\caption{\textit{\textbf{Dataset description.} The data was collected across four distinct environments: (i) a laboratory with minimal background clutter, (ii) a dining area featuring moderate background clutter, (iii) a kitchen with a similar level of background clutter, and (iv) an outdoor garden characterized by high background clutter. Different acquisitions are done in each environment.}}
\begin{tabularx}{0.47\textwidth}{@{}l*{2}{c}c@{}}
\toprule
Environment  & Nbr. of Acquisitions &  Nbr. of Frames &  Difficulty  \\ 
\midrule
Laboratory   &     4     &  12114 &  Low      
\\ 
Dining place   &    4      &  12130  &  Medium
\\ 
Kitchen   &     4     &  12124    &  Medium    
\\
Garden  &    6     &   15193   &  High      
\\
\bottomrule
\end{tabularx}
\label{datasettable}
%\vspace{-12pt}
\end{table}

In the next Section, we describe our CNN architecture for the inference of rotational speed $y$ from the thermal camera data $\Tilde{X}$ which will be trained using the dataset of Table \ref{datasettable}.

\section{CNN Architecture}
\label{cnnsetup}

The CNN fusion architecture employed in this study is illustrated in Fig. \ref{cnnarch}. It has been specifically designed to maintain a compact structure, minimizing memory usage and computational complexity for efficient deployment on CNN accelerator hardware, such as Google’s Coral Edge TPU \cite{googlecoraltpu}. The CNN fusion model in Fig. \ref{cnnarch} is a modular architecture composed of \textit{i)} a thermal feature extraction backbone; \textit{ii)} a thermal-only odometry estimation readout (estimating the rotation speed using thermal data only); \textit{iii)} a second readout which infers the fusion gain parameter $K_g$; and the thermal-gyroscope fusion block which fuses the thermal-only odometry estimate with the average gyroscope readings using the inferred fusion gain parameter $K_g$.

\begin{figure}[htbp]
%\vspace{-15pt} 
\centering
    \includegraphics[scale = 0.55]{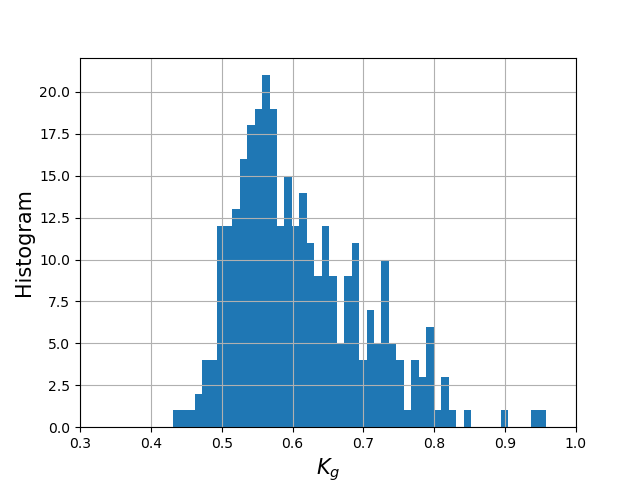}
    \caption{\textit{\textbf{Histogram of the fusion gain $K_g$ recorded over $6$ CNN inference cycles (using six different input data sequences):.} It can be seen that the value of $K_g$ is distributed around a peak at $K_g=0.55$. The value of $K_g$ is inferred by the CNN depending on the thermal camera quality and balances the fusion between the gyro readings and the rotation speed estimated from the thermal data.  }}
    \label{kgress}
    %\vspace{-10pt}
\end{figure}

The thermal feature extraction backbone features a first convolutional layer $\text{Conv}_\text{1}$ with $6$ filters of size $5\times5$. Then, the output tensor of $\text{Conv}_\text{1}$ is fed to a max pooling layer with $\times 2$ down sampling. After maxpooling, a second convolutional layer $\text{Conv}_\text{2}$ is used with $16$ filters of size $5\times5$. Regarding the thermal-only odometry readout, the output of the $\text{Conv}_\text{2}$ layer is flatten and fed to two fully-connected layers $\text{FC}_\text{1}$ and $\text{FC}_\text{2}$ with size $120$ and $80$ neurons, before being processed by a linear layer producing a scalar (1D) output $\hat{y}_{th}$ estimating the rotational speed using the thermal camera data only.

Regarding the inference of the fusion gain parameter $K_g$, the output of the backbone is flatten and fed to a fully-connected layer with $120$ ReLU neurons followed by a single \textit{sigmoid} neuron output which provides the fusion gain $K_g$ such that $0\leq K_g \leq 1$.

Then, $K_g$ is used to adaptively fuse the rotation speed $\hat{y}_{th}$ estimated via the thermal camera together with the rotation speed $\hat{y}_{gy}$ estimated by locally averaging the gyroscope readings (where $g_l$ denotes the gyroscope output of time index $l$ and $N_f$ is the consecutive number of frames fed into the network):
\begin{equation}
    \hat{y}_{gy} = \frac{1}{N_f} \sum_{l=1}^{N_f} g_l
\end{equation}
The \textit{final fusion output} is derived as follows:
\begin{equation}
    \hat{y}_{fus} = K_g \times \hat{y}_{th} + (1 - K_g) \times \hat{y}_{gy}
\end{equation}
and is inspired by how sensor fusion is achieved in Kalman filters \cite{kalmanfilter, deepkalman}, where $K_g$ acts as a \textit{learnable} Kalman gain (see Fig. \ref{kgress}) that is adaptively inferred by the CNN itself depending on the statistics of the input thermal camera data (the noisier the thermal data, the smaller $K_g$ would be to put more weight on the gyroscope reading, and vice versa).

The network receives a sequence of $N_f$ consecutive thermal camera frames as input. The CNN architecture, illustrated in Figure \ref{cnnarch}, employs ReLU activation functions and is optimized using the Adam optimizer \cite{adampaper} with a learning rate of $\eta = 0.001$ and a batch size of $B = 32$ over 40 epochs. The loss function $L$ is defined as the \textit{inverted Huber loss} \cite{berhu}, which quantifies the error between the CNN fusion output $\hat{y}_{fus}$ and the ground truth rotational speed $y$.

\begin{equation}
    \mathcal{L} = \begin{cases} |\hat{y}_i - y_i|, & \mbox{if } |\hat{y}_i - y_i| \leq c \\ \frac{(\hat{y}_i - y_i)^2 + c^2}{2c}, & \mbox{else}\end{cases}
    \label{berhu}
\end{equation}

This choice of loss function is motivated by the fact that the inverted berHu loss (\ref{berhu}) puts more emphasis on the difficult examples during training (corresponding to the quadratic region $|\hat{y}_i - y_i| > c$ in (\ref{berhu})) \cite{berhu}. Similar to \cite{sensorfusion}, we adaptively set the $c$ parameter of (\ref{berhu}) as $c = 0.2 \times \max_{i} |\hat{y}_i - y_i|$ where the $i$ index denotes the $i^{th}$ element in the mini batch of training labels. During our experiments, we observed that using the inverted Huber loss always led to a higher test precision compared to the use of the conventional \textit{mean square error} (MSE) loss \cite{mse}, further motivating the use of (\ref{berhu}).
%\begin{equation}
%    \mathcal{L} = \frac{1}{B} \sum_{i = 1}^{B} ||\hat{y}_i - y_i ||_2^2
%\end{equation}

In the following section, we analyze the effect of the number of consecutive frames $N_f$ on the CNN inference precision. Additionally, we examine the impact of the thermal camera \textit{resolution} subsampling factor $N_r$ by progressively downsampling the input frames to assess its influence on CNN performance. Understanding the extent to which input signal dimensionality can be reduced will, in turn, enable a decrease in overall memory consumption and computational complexity of the proposed CNN-based system for hardware implementation.

%Through these experiments, our aim is to study how much the input signal dimensionality can be reduced, reducing in turn the overall memory consumption and compute complexity of the proposed CNN-based system. 

%Experiments and 
\section{Results}
\label{resultssection}

\color{black}

The objective of the experimental investigations is to evaluate the performance of our thermal-gyro fusion-based rotational odometry system and compare it to the thermal-only system under varying configurations of two key parameters: the number of consecutive frames (\(N_f\)) provided as input to the CNN and the resolution of the thermal camera images (\(N_r\)). Understanding the impact of \(N_f\) and \(N_r\) on the precision of the odometer is critical, as these parameters directly influence the dimensionality of the input data, subsequently affecting the computational complexity of the model and the memory requirements.  Thus, reducing these parameters while maintaining a tolerable level of performance results in a more efficient implementation for resource-constrained on-chip CNN systems, making it suitable for hardware deployment in edge devices.

\subsection{Impact of the number of consecutive frames $N_f$}
\label{nfimpact}
\begin{figure}[htbp]
\centering
%\vspace{-20pt} 
    \includegraphics[scale = 0.55]{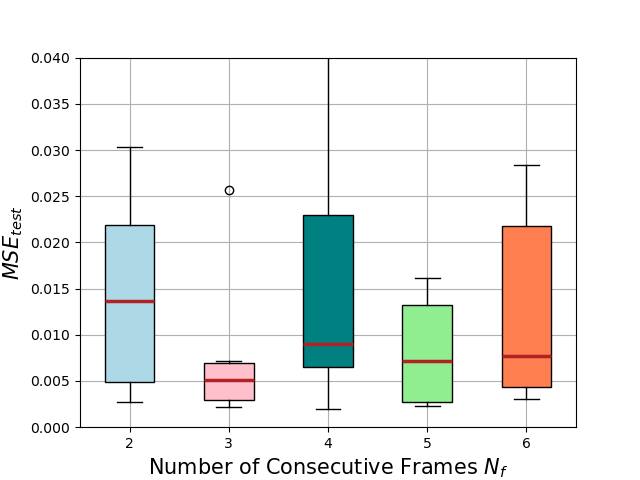}
    \caption{\textit{\textbf{Thermal-only case: Box plot of the 6-fold test MSE in function of the number of consecutive thermal input frames $N_f$.} The red line indicates the median value. The best $\text{MSE}_{\text{test}}$ is achieved for $N_f=3$. }}
    \label{res1}
%    \vspace{-13pt}
\end{figure}

\begin{figure}[htbp]
    %\vspace{-10pt}
\centering
    \includegraphics[scale = 0.55]{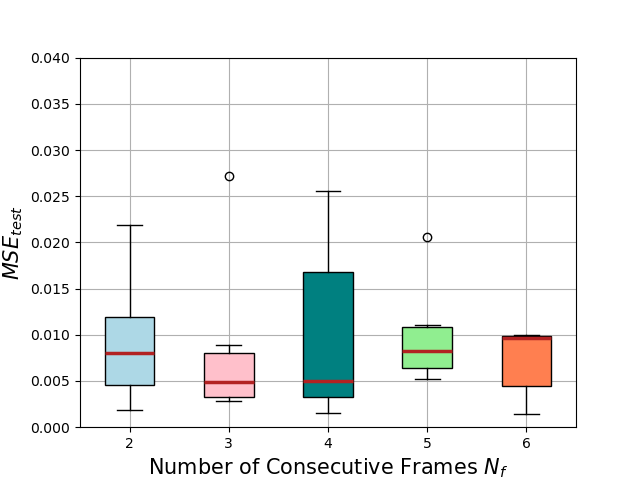}
    \caption{\textit{\textbf{Thermal-Gyro fusion case: Box plot of the 6-fold test MSE in function of the number of consecutive thermal input frames $N_f$.} The red line indicates the median value. The best $\text{MSE}_{\text{test}}$ is achieved for $N_f=3$. }}
    \label{res1_fus}
    %\vspace{-20pt}
\end{figure}

We investigate the effect of the number of consecutive frames $N_f$ on the test precision of the CNN, as shown in Figure  \ref{cnnarch}, under both \textit{thermal-only} and \textit{thermal-gyro} fusion setups. To assess this impact, we employ a systematic 6-fold train-test procedure. In each iteration, one of the six independent acquisitions from the challenging Garden environment (as described in Table~\ref{datasettable}) was selected as the test set, while the remaining acquisitions were used for training following the methodology outlined in Section~\ref{cnnsetup}. This procedure was repeated for every acquisition, and the final test mean squared error (\(\text{MSE}_{\text{test}}\)) for different values of \(N_f\) (ranging from 2 to 6) is presented as box plots in Fig. ~\ref{res1} (thermal-only) and Fig. ~\ref{res1_fus} (thermal–gyro fusion). Both configurations demonstrate that \(N_f = 3\) yields the lowest \(\text{MSE}_{\text{test}}\). The trend observed in Fig. \ref{res1} can be interpreted as follows: when $N_f = 2$, the CNN receives insufficient input data, resulting in under-fitting and a high $MSE_{\text{test}}$. Conversely, for $N_f > 3$, the model processes excessive input frames, introducing redundant data that increases the risk of overfitting. 

\begin{figure}[htbp]
%\vspace{-20pt}
\centering
    \includegraphics[scale = 0.55]{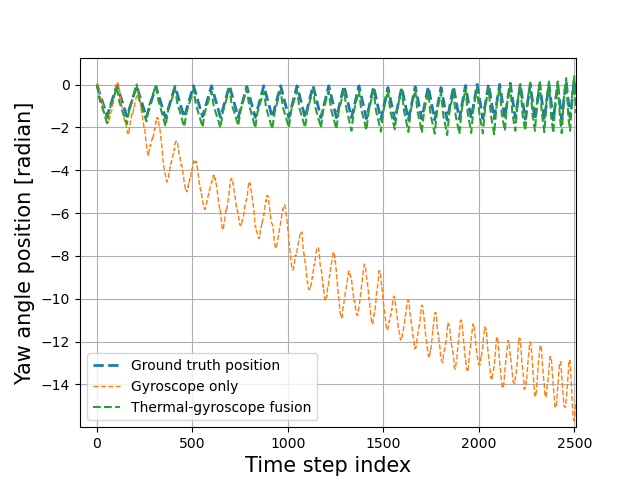}
    \caption{\textit{\textbf{Angular position estimation using the proposed thermal-gyro fusion vs. gyro-only.} The angular position is estimated by integrating the angular speed readings through time. Due to the well-known biases present in MEMS gyroscope sensor chips \cite{biasgyro}, the angular position estimation using the gyroscope readings drifts away during the integration. On the other hand, the position estimation using the proposed thermal-gyroscope fusion does not suffer from drifting during the integration.  }}
    \label{comparison_imu}
    %\vspace{-20pt}
\end{figure}

In particular, while both configurations achieve their lowest error at
\(N_f = 3\), Table~\ref{table1} reveals that the fusion setup consistently achieves lower median errors and, more importantly, a reduced interquartile range (IQR) compared to the thermal-only setup across the $N_f$ values. The smaller IQR in the fusion case indicates more \textit{consistent} errors across different test sequences, which is crucial for ensuring reliable performance in real-world applications. Moreover, as shown in Fig.~\ref{comparison_imu}, our thermal-gyro fusion approach effectively mitigates the well-known drift limitation of gyroscope-based odometry. Specifically, Fig.~\ref{comparison_imu} illustrates that with
$N_f = 3$, gyro-only integration results in increasing deviation from the ground truth over time, while the proposed fusion method maintains stable angular position estimates.

% During our experiments, $N_f$ is swept from $2$ to $6$. When using the thermal camera data only, Fig. \ref{res1} shows that the lowest $\text{MSE}_{\text{test}}$ is achieved for $N_f = 3$. The trend in Fig. \ref{res1} can be explained as follows: for $N_f=2$, the CNN receives too little input data and \textit{under-fits}, leading to a high $\text{MSE}_{\text{test}}$. For values of $N_f>3$, the CNN receives an excessively large amount of input frames, leading to potential \textit{over-fitting}. On the other hand, $N_f=3$ leads to the best CNN fitting performance. 


% \textcolor{blue}{make link to Table 2 in the text and explain that the fusion setup clearly achieves better performances in terms of median error but more importantly in terms of IQR which denotes how much "variance" is in the results (the smaller IQR the more the errors are similar and not very different depending on different test data sequences}

% \textcolor{orange}{also make link with Figure 6 to clearly convince the readers that our approach does not suffer from the famous problem of estimation drift encountered when integrating the output of gyroscope through time, this should be enough to convince the readers about the usefulness of our approach vs. gyro only}

\begin{table}[htbp]
\centering
\caption{\textit{\textbf{Median error and Inter Quartile Range (IQR) in function of the number of input frames $N_f$.} The thermal-gyro fusion and thermal-only setup are respectively indicated by (fus) and (no fus). }}
\begin{tabularx}{0.47\textwidth}{@{}l*{3}{c}c@{}}
\toprule
$N_f$  & Median \textbf{(no fus)} & Median \textbf{(fus)} & IQR \textbf{(no fus)} & IQR \textbf{(fus)}  \\ 
\midrule
%HOTS \cite{7508476}      & 80.8       & 27.1          \\ 
2      &  0.014    &  \textbf{0.008} &  0.017 & \textbf{0.007}     \\ 
3      &  \textbf{0.005}  & \textbf{0.005} &  \textbf{0.004} & 0.005  \\ 
4    &    0.009   & \textbf{0.005} &  0.016 & \textbf{0.014} \\ 
5    &  \textbf{0.007}  & 0.008  &  0.011 & \textbf{0.005}  \\ 
6    &  \textbf{0.008}  & 0.009 &  0.018 & \textbf{0.005}    \\

%\midrule
%[0.00734676 0.00472887 0.01351124 0.00447449 0.00543081]
%NeuNorm \cite{DBLP:journals/corr/abs-1809-05793} &  99.53 & 60.5  \\
%Deep STS-ResNet \cite{DBLP:journals/corr/abs-2003-12346} &  99.6 & 69.2  \\
\bottomrule
\end{tabularx}

\label{table1}

\end{table}

Since $N_f=3$ yields the best performance in both Figures \ref{res1} and \ref{res1_fus}, we will use this setting in Section \ref{resol} to examine the impact of thermal camera resolution on CNN precision under both the thermal-gyro fusion and thermal-only configurations.

\subsection{Impact of the thermal camera resolution $N_r$}
\label{resol}

\begin{figure}[htbp]
%\vspace{-25pt}
\centering
    \includegraphics[scale = 0.55]{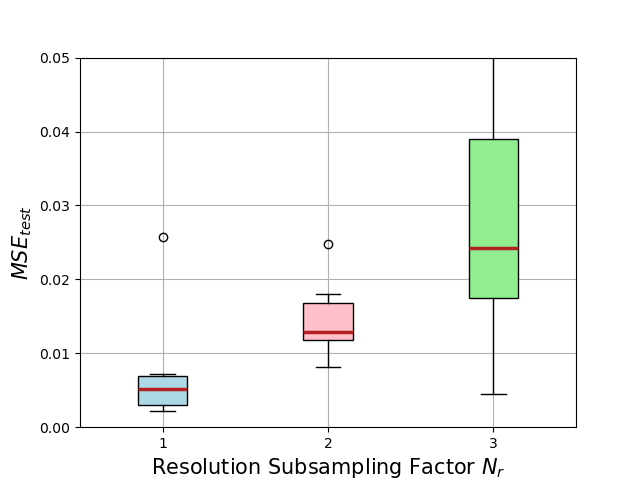}
    \caption{\textit{\textbf{Thermal-only case: Box plot of the 6-fold test MSE in function of the thermal camera resolution subsampling factor $N_r$.} The red line indicates the median value. As expected, the lower the thermal image resolution, the higher the $\text{MSE}_{\text{test}}$. }}
    \label{res2}
    %\vspace{-6pt}
\end{figure}

\begin{figure}[htbp]
%\vspace{-20pt}
\centering
    \includegraphics[scale = 0.55]{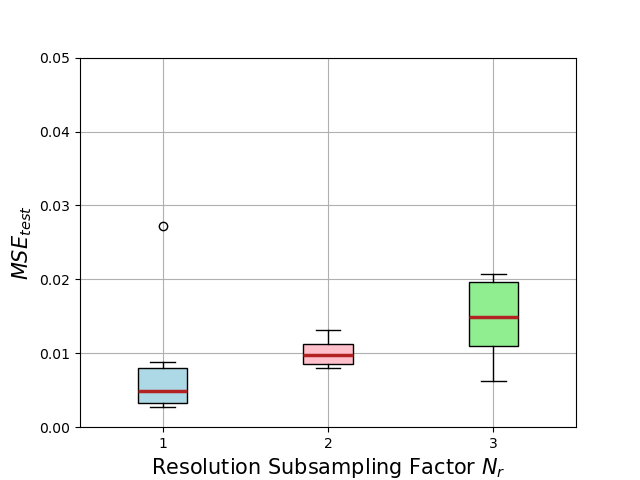}
    \caption{\textit{\textbf{Thermal-gyro fusion case: Box plot of the 6-fold test MSE in function of the thermal camera resolution subsampling factor $N_r$.} The red line indicates the median value. As expected, the lower the thermal image resolution, the higher the $\text{MSE}_{\text{test}}$. }}
    \label{res2_fus}
    %\vspace{-20pt}
\end{figure}

We now investigate the impact of the thermal camera resolution subsampling factor \(N_r\) on the CNN test precision and computational efficiency. Following the same 6-fold train-test procedure, we report the box plots of \(\text{MSE}_{\text{test}}\) as a function of \(N_r\) for both thermal-only (Fig.~\ref{res2}) and thermal-gyro fusion (Fig.~\ref{res2_fus}) configurations. The subsampling factor \(N_r\) is varied across \(\{1, 2, 3\}\), where higher values indicate greater downsampling, achieved by locally averaging neighboring pixels in the thermal frames. As expected, reducing the thermal image resolution leads to higher \(\text{MSE}_{\text{test}}\), with the most pronounced degradation observed for \(N_r = 3\). This trend is consistent in both setups, confirming that spatial resolution is critical for accurate rotation estimation. However, as shown in Table \ref{table2}, the fusion setup consistently performs better with $30\%$ lower median errors on average compared to the thermal-only setup, even as subsampling increases. Crucially, fusion clearly achieves a significantly reduced IQR again (average reduction of $92\%$ across the $N_r$ values), indicating that the errors remain more consistent across different test sequences, which is crucial for robust odometry estimation.


 A key takeaway from these results is that \textit{thermal-gyro fusion allows the system to operate at lower thermal resolutions while not compromising accuracy and IQR as significantly as in the thermal-only case}, addressing a key limitation of thermal-only odometry. In the thermal-only setup, reducing resolution significantly increases \(\text{MSE}_{\text{test}}\) and introduces substantial error variability (higher IQR). In contrast, fusion mitigates this effect, enabling further resolution reduction without sacrificing precision.
 \begin{table}[htbp]
\centering
\caption{\textit{\textbf{Median error and Inter Quartile Range (IQR) in function of the thermal camera resolution subsampling factor $N_r$.} The thermal-gyro fusion and thermal-only setup are respectively indicated by (fus) and (no fus). }}
\begin{tabularx}{0.47\textwidth}{@{}l*{3}{c}c@{}}
\toprule
$N_r$  & Median \textbf{(no fus)} & Median \textbf{(fus)} & IQR \textbf{(no fus)} & IQR \textbf{(fus)}  \\ 
\midrule
%HOTS \cite{7508476}      & 80.8       & 27.1          \\ 
1      &  \textbf{0.005}    &  \textbf{0.005} &  \textbf{0.004} & 0.005   \\ 
2     &  0.013  & \textbf{0.01} &  0.005 & \textbf{0.003}  \\ 
3    &    0.024   & \textbf{0.015} &  0.022 & \textbf{0.009} \\ 

%\midrule
%[0.00734676 0.00472887 0.01351124 0.00447449 0.00543081]
%NeuNorm \cite{DBLP:journals/corr/abs-1809-05793} &  99.53 & 60.5  \\
%Deep STS-ResNet \cite{DBLP:journals/corr/abs-2003-12346} &  99.6 & 69.2  \\
\bottomrule
\end{tabularx}

\label{table2}

\end{table}
 
\begin{figure}[H]
%\vspace{-10pt}
\centering
    \includegraphics[scale = 0.5]{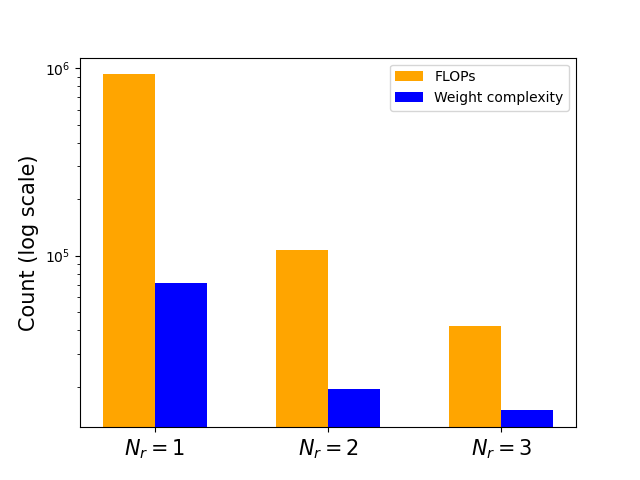}
    \caption{\textit{\textbf{CNN compute complexity and weight parameter complexity in function of the thermal camera resolution subsampling factor $N_r$.} The higher the subsampling factor $N_r$, the lower the number of Floating Point Operations (FLOPs) and the number of CNN weight parameters, reducing both compute complexity and memory consumption.}}
    \label{computecomp}
    %\vspace{-15pt}
\end{figure}

 This advantage is particularly crucial for computational efficiency, as lower-resolution inputs reduce CNN complexity, memory requirements, and energy consumption. As shown in Fig.~\ref{computecomp}, increasing  \(N_r\) leads to a substantial decrease in Floating Point Operations (FLOPs) and weight complexity, resulting in lower power consumption, faster inference times, and reduced hardware demands—critical factors for edge deployment in resource-constrained compute platforms. 
Many small mobile robots, particularly those with limited onboard computing power and battery capacity, struggle with high-resolution image processing due to energy constraints. By leveraging fusion, computational load is significantly reduced (see Fig. \ref{computecomp}), making this system highly suitable for real-time deployment in power-efficient robotic applications.

\section{Conclusion}
\label{concsection}
This study presented a novel thermal-gyro fusion approach for rotational odometry, integrating ultra-low-resolution thermal imaging with gyroscope data. Through systematic evaluation, we demonstrated that this fusion enables a significant reduction in thermal camera resolution without significantly compromising accuracy, offering a compute-efficient and cost-effective solution for real-time deployment on resource-constrained robotic platforms. Our experimental analysis provided key insights into optimal sensor fusion configurations, revealing the influence of thermal frame count and resolution on odometry performance. Finally, we have openly released our thermal-gyro fusion dataset as supplementary material with the hope of benefiting future research. As future work, we plan to extend our system beyond rotational odometry only, towards the estimation of translational odometry as well, by fusing the thermal camera with accelerometer data. %Notably, our method maintains stable rotational estimates while significantly reducing memory and computational demands, making it particularly well-suited for edge AI and embedded robotic applications. Furthermore, the results highlight the robustness of our approach across varying environmental conditions, positioning it as a viable alternative to traditional visual-inertial odometry systems, particularly in low-light or visually degraded settings. To foster further research in low-cost, sensor fusion-based odometry, we publicly release our multimodal dataset, paving the way for enhanced autonomy in robotics and related fields.

%%%%%%%%%%%%%%%%%%%%%%%%%%%%%%%%%%%%%%%%%%%%%%%%%%%%%%%%%%%%%%%%%%%%%%%%%%%%%%%%

% The research leading to these results has received funding from the Flemish Government (AI Research Program) and the European Union's ECSEL Joint Undertaking under grant agreement n° 826655 - project TEMPO.

%%%%%%%%
%MAP1: RGB: map 0.1572, loc 0.588; orig: map: 0.22, loc 0.72; VGG map: 1.92, loc 3.95
%MAP2: RGB: map 0.329, loc 0.742; orig: map: 0.248, loc 1.499; VGG map: 1.96, loc 2.58
%MAP3: RGB: map 0.244, loc 1.596; orig: map: 1.15, loc 2.714; VGG map: 0.266; loc 1.775
%%%%%%%%

\end{document}